%% file: arxiv.tex
\documentclass{article}
\usepackage[letterpaper, total={6in, 8in}]{geometry}

\usepackage{authblk}

\RequirePackage{graphicx}
\RequirePackage{amsfonts,amssymb,amsbsy,amsmath,amsthm}

\input{preamble.tex}
\begin{document}

\title{A Brief Survey on the Approximation Theory for \\Sequence Modelling}

%
%
%
\author[1]{
  Haotian Jiang
}
\author[1,2]{
  Qianxiao Li
  \thanks{
    Corresponding author.
    {\tt qianxiao@nus.edu.sg}.
  }
}
\author[3]{
  Zhong Li
}
\author[1]{
  Shida Wang
}
\affil[1]{Department of Mathematics, National University of Singapore}
\affil[2]{Institute for Functional Intelligent Materials, National University of Singapore}
\affil[3]{Microsoft Research Asia}



%
%
%

\date{}
\maketitle

\begin{abstract}
  We survey current developments in the approximation theory of sequence modelling in machine learning.  Particular emphasis is placed on classifying existing results for various model architectures through the lens of classical approximation paradigms, and the insights one can gain from these results.  We also outline some future research directions towards building a theory of sequence modelling.
\end{abstract}
\input{intro.tex}

\input{formulation.tex}

\input{rnn.tex}

\input{others.tex}

\input{outlook.tex}


\section*{Acknowledgments}

We thank the reviewers for their comments,
and Lukas Gonon, Lyudmila Grigoryeva and
Juan-Pablo Ortega for their useful suggestions
on the manuscript.

QL is supported by the National Research Foundation, Singapore,
under the NRF fellowship (project No. NRF-NRFF13-2021-0005).
HJ is supported by National University of Singapore under the PGF scholarship.
SW is supported by National University of Singapore under the Research scholarship.

\bibliography{library}

\appendix

\end{document}

%% file: preamble.tex
\usepackage{caption}
\usepackage{subcaption}
\usepackage{xcolor}
\usepackage{graphicx}
\usepackage[final]{hyperref}
\usepackage[numbers]{natbib}
\usepackage{bm}
\usepackage{float}
\usepackage[nameinlink,capitalize]{cleveref}
\usepackage{booktabs}

\newcommand{\R}{\mathbb{R}}
\newcommand{\Z}{\mathbb{Z}}
\newcommand{\N}{\mathbb{N}}
\newcommand{\E}{\mathbb{E}}
\newcommand{\Conv}{\mathop{\scalebox{1.7}{\raisebox{-0.2ex}{$\ast$}}}}%

\newcommand{\seq}[1]{\boldsymbol{#1}}
\newcommand{\set}[1]{\mathcal{#1}}
\newcommand{\h}[1]{\widehat{#1}}
\newcommand{\Hrnn}{\set{H}_{\text{RNN}}}
\newcommand{\Hrnnl}{\set{H}_{\text{L-RNN}}}
\newcommand{\Hcnn}{\set{H}_{\text{CNN}}}
\newcommand{\Hcnnl}{\set{H}_{\text{L-CNN}}}
\newcommand{\Hesn}{\set{H}_{\text{ESN}}}
\newcommand{\Htrans}{\set{H}_{\text{Trans}}}
\newcommand{\Htransp}{\set{H}_{\text{SpTrans}}}
\newcommand{\fg}{hidden dynamic functional sequences}
\newcommand{\Cfg}{\set{C}_{HD}}
\newcommand{\Cfmp}{\set{C}_{\text{FMP}}}
\newcommand{\normw}[1]{\|#1\|_{\seq{w}}}
\newcommand{\rencdec}{{REncDec}}

\newcommand{\Hedl}{\set{H}_{\text{L-\rencdec}}}
\newcommand{\Crnn}{\set{C}_{\text{RNN}}}
\newcommand{\normCrnn}[1]{\|#1\|_{\text{RNN}}}
\newcommand{\tens}[1]{\mathbf{T}(#1)}

\newcommand{\rev}[1]{{#1}}
\newcommand{\revtwo}[1]{{#1}}

%% file: intro.tex
\section{Introduction}
\label{sec:intro}


The modelling of relationships between sequences is an important task that enables
a wide array of applications, including
classical time-series prediction problems in finance~\citep{taylor2008.ModellingFinancialTime},
and modern machine learning problems in natural language processing~\citep{bahdanau2016.NeuralMachineTranslation}.
Another class of engineering applications involving sequential relationships are control systems,
which study the dependence of a dynamical trajectory on an input
control sequence~\citep{brunton2019.DatadrivenScienceEngineering}.
In general, sequence-to-sequence relationships can be very complex.
For example, when the index set for the sequences is infinite,
one can understand these relationships as mappings between infinite-dimensional spaces.
Thus, traditional modelling techniques are limited in their efficacy,
especially when there is little prior knowledge on the system of interest.
To address these difficulties,
an increasingly popular method to model sequence relationships is to
leverage machine learning.

To date, a large variety of machine learning paradigms have been proposed to
model sequential relationships.
One of the earliest attempts is the class of neural networks called
\emph{recurrent neural networks (RNN)}%
~\citep{rumelhart1986.LearningRepresentationsBackpropagating},
and their variants~\citep{cho2014.LearningPhraseRepresentations,hochreiter1997.LongShortTermMemory}.
Besides the RNN family of models, many other alternatives have also been explored.
These include convolutional based models~\citep{oord2016.WaveNetGenerativeModela},
encoder-decoder based models~\citep{cho2014.LearningPhraseRepresentations}
attention based models~\citep{bahdanau2016.NeuralMachineTranslation},
and their combinations.
For example, the powerful transformer architecture~\citep{vaswani2017.AttentionAllYou}
combines encoder-decoder and attention architectures.

Despite the rapid developments in the practical domain,
the theoretical foundation of data-driven sequence modelling is still in its nascent stages.
For example, the most basic question of how the aforementioned architectures are different,
and how practitioners should select the model architecture based on their applications,
is largely unknown and relies on trial and error.
Thus, an important direction of theoretical research is to understand the essential properties,
and most importantly, distinctions between different sequence modelling paradigms.

The present survey aims to provide an overview of the theoretical research
on sequence modelling in the specific direction of \emph{approximation theory}
\citep{lorentz2005.ApproximationFunctions,devore1998.NonlinearApproximation}.
In a nutshell, approximation theory is the study of how a complex relationship (say a function)
can be broken down as a combination of simpler, more readily computable building blocks.
A theoretical foundation of sequence modelling requires the understanding of how and when
a sequential relationship can be approximated by simpler components realized as various
neural network architectures.
The theory of sequence modelling is an active area of research that spans decades
of work both in machine learning and the study of nonlinear dynamics.
Thus, the purpose of this survey is not to give an exhaustive summary of all relevant results in the literature,
but rather to highlight some interesting insights for approximation theory gained from existing works.
Moreover, we discuss some classes of open questions that are of significance in order
to progress the understanding of the approximation theory for sequences.

The survey is organized as follows.
In~\cref{sec:formulation}, we introduce the mathematical problem of approximation,
including the key questions one is interested in answering.
In particular, we highlight the new aspects of sequence approximation as compared to classical paradigms
of function approximation.
In~\cref{sec:rnn}, we discuss approximation results on recurrent neural networks, where much more is known
compared with other architectures.
In~\cref{sec:others}, we consider the approximation theory of other model architectures, including those of convolutional,
encoder-decoder and attention types.
In~\cref{sec:outlook}, we summarize the known results and motivate some future directions of interest.

\paragraph{Notation.}
Let us introduce some notational conventions.
Throughout this paper, we use lower-case letters to denote scalars and vectors.
Boldface letters are reserved for sequences, e.g. $\seq{x} = \{ x(t) : t\in \set{T} \}$.
As in the previous formula, script letters such as $\set{T}$
are used to represent sets of scalar or vectors, both finite and infinite-dimensional.
Capital letters are used to denote mappings between vector spaces.
Correspondingly, a bold-faced capital letter is a sequence of such mappings.
Sometimes, we wish to refer to a portion of a sequence.
Let $\set{S} \subset \set{T}$,
then $\seq{x}_{\set{S}} := \{x(t) : t \in \set{S}\}$.
We use $|\cdot|$ to denote the Euclidean norm,
and reserve $\| \cdot \|$, possibly with subscripts, for
norms over function (e.g. sequence) spaces.
We use $\dot{x}(t)$ to denote the derivative of $t \mapsto x(t)$.
Higher derivatives of order $r\geq 0$ are written as $x^{(r)}(t)$.
Throughout this survey, we reserve the letters $m,n,d,i,j$
to represent integers.
Sequence indices, equivalently referred to as time indices,
are written as $t$ or $s$.

%% file: formulation.tex
\section{Sequence modelling as an approximation problem}
\label{sec:formulation}


We begin by formalizing the broad mathematical problem of approximation and some of the key questions
one may be interested in.
We then discuss how one may formulate sequence modelling
in the setting of approximation theory.

\subsection{The problem of approximation}
\label{subsec:formulation_approx}

Let us introduce the basic problem of approximation for functions on vector spaces.
Let $\set{X}$ and $\set{Y}$ be normed vector spaces.
We consider a family of \emph{target functions}, or simply \emph{targets},
which is a subset $\set{C}$ of all mappings $\set{X} \to \set{Y}$,
i.e. $\set{C} \subset \set{Y}^\set{X}$.
In the learning theory literature, one sometimes calls $\set{C}$ a \emph{concept space}.
These are the relationships we wish to learn, or approximate, by some
simpler candidate functions.
Let us denote this set of candidates by $\set{H} \subset \set{Y}^\set{X}$.
In learning theory, this is often called a \emph{hypothesis space}.
The problem of approximation concerns how well can functions in $\set{H}$ resolve
functions in $\set{C}$.

In broad terms, we may classify results on approximation theory into three types:
\emph{universal approximation results (density-type)},
\emph{approximation rate estimates (Jackson-type)},
and \emph{inverse approximation results (Bernstein-type)}.
Let us discuss each in turn.

\paragraph{Universal approximation results (density-type).}

Universal approximation theorems are the most basic approximation results.
We say that $\set{H}$ is an universal approximator for $\set{C}$ if
for every $H \in \set{C}$ and $\epsilon > 0$,
there exists $\h{H} \in \set{H}$ such that $\| H - \h{H} \| \leq \epsilon$.
In other words, $\set{H}$ is dense in $\set{C}$ in the topology generated
by $\| \cdot \|$.
The choice of the norm depends on applications.
We illustrate this with the following example.

We consider approximating scalar functions by trigonometric polynomials.
Here, we set $\set{X} = [0, 2\pi]$ and $\set{Y} = \R$.
The target space is $\set{C} = C^{\alpha}_{\text{per}}([0,2\pi])$,
the set of $\alpha$-times continuously differentiable, periodic functions on $[0,2\pi]$.
The hypothesis space is
\begin{equation}\label{eq:H_trig}
    \set{H} =
    \cup_{m\in \N_+}
    \left\{
        \h{H}(x) =
        \sum_{i=0}^{m-1}
        a_i \cos(i x)
        +
        b_i \sin(i x)
        :
        a_i, b_i \in \R,
        m \in \N_+
    \right\}.
\end{equation}
As a direct consequence of the Stone-Weierstrass theorem,
$\set{H} \subset \set{C}$ is dense in $\set{C}$ with respect to the norm
$\| H \| = \sup_{x\in[0,2\pi]} |H (x)|$
\citep[p.~32]{achieser2013.TheoryApproximation}.

\paragraph{Approximation rate estimates (Jackson-type).}
Universal approximation (density) ensures that our hypothesis space $\set{H}$
is in a sense ``big enough'', so that we can use it to approximate a reasonably large
variety of target functions.
However, such results do not tell us precisely what types of functions in $\set{C}$
are ``easy'' (or ``hard'') to approximate using $\set{H}$.
In other words, two hypothesis spaces $\set{H}_1$ and $\set{H}_2$ may both be dense in
$\set{C}$ but can be naturally adapted to approximate functions of different types.

To resolve this, we may ask a finer question on the rate of approximation.
Fix a hypothesis space $\set{H}$.
Let $\{ \set{H}^m : m \in \N_+ \}$ be a collection of subsets of $\set{H}$ such that
$\set{H}^m \subset \set{H}^{m+1}$ and $\cup_{m \in \mathbb{N}_+} \set{H}^m = \set{H}$.
Here, $m$ is a measure of complexity of the approximation candidates,
and $\set{H}^m$ is the subset of hypotheses with complexity at most $m$.
This is also called the approximation budget.
Then, the approximation rate estimate is an inequality of the form
\begin{equation}\label{eq:qn_jackson_type}
    \inf_{\h{H} \in \set{H}^m}
    \| H - \h{H} \|
    \leq
    C_{\set{H}}(H, m).
\end{equation}
\cref{eq:qn_jackson_type} tells us the best possible approximation error
one can hope to obtain under approximation budget $m$.
Note that $\set{H}$ is dense if and only if $\lim_{m\to\infty} C_{\set{H}}(H,m) = 0$
for every $H \in \set{C}$.
The speed at which $C_{\set{H}}(H,m)$ decays as $m$ increases is the approximation rate,
and its dependence on $H$ measures the complexity of a particular target $H$
under the current approximation scheme $\set{H}$.

Returning to the example in~\cref{eq:H_trig}, the hypothesis space with budget $m$ is
\begin{equation}
    \set{H}^m =
    \left\{
        \h{H}(x) =
        \sum_{i=0}^{m-1}
        a_i \cos(i x)
        +
        b_i \sin(i x)
        :
        a_i, b_i \in \R
    \right\}.
\end{equation}
    The classical Jackson's theorem~\citep[p.~187]{achieser2013.TheoryApproximation}
    gives a rate estimate of the form

\begin{equation}\label{eq:jackson_trig}
    \inf_{\h{H} \in \set{H}^m}
    \| H - \h{H} \|
    \leq
    \frac
    {
        c_\alpha
        \max_{0\leq r\leq \alpha}
        \| H^{(r)} \|
    }
    {
        m^{\alpha}
    },
\end{equation}
where $c_\alpha$ is a constant depending only on $\alpha$.
Observe that the rate of decay of the approximation error is $m^{-\alpha}$
and the complexity of a target function under the trigonometric polynomial approximation scheme
is its norm associated with the Sobolev space
$\mathcal{W}^{\alpha,\infty} := \{ H: \max_{0\leq r\leq \alpha} \| H^{(r)} \| < \infty \}$.
The key insight here is that a function $H$ is easy to approximate using trigonometric polynomials
if it has small gradient (Sobolev) norm.
We will hereafter refer to estimates in the form of~\cref{eq:qn_jackson_type} as
\emph{Jackson-type} results.

\paragraph{Inverse approximation results (Bernstein-type).}
Jackson-type results tell us that if a target function $H$ possesses some property related to $\{\set{H}^m\}$,
(e.g. smoothness, small gradient norm), then it is in fact easy to approximate with $\{\set{H}^m\}$.
Inverse approximation results are converse statements.
It identifies properties that $H$ ought to possess if one starts with the assumption that it
can be well-approximated (in a sense to be made precise in each case) by $\{ \set{H}^m \}$.

In the case of trigonometric polynomial approximation,
the following inverse approximation result is due to Bernstein~\citep[p.~206]{achieser2013.TheoryApproximation}.
Fix some periodic $H:[0,2\pi]\to \R$, and suppose that there exists
a constant $c>0$, $\delta >0$ and $\alpha \in \N_+$
so that for every $m\in \N_+$, one has

\begin{equation}\label{eq:berstein_trig}
    \inf_{\h{H} \in \set{H}^m}
    \| H - \h{H} \| \leq
    \frac
    {c}
    {m^{\alpha + \delta}}.
\end{equation}
Then, $H$ is $\alpha$-times continuously differentiable and its $\alpha$-th
derivative is $\delta$-H\"{o}lder continuous.
Intuitively, this result says that if a function $H$ can be approximated with a rate
in~\cref{eq:jackson_trig}, then it must be in $C^\alpha_{\text{per}}([0,2\pi])$.
Combined with Jackson's result, one gains a complete characterization of the type of functions
- namely smooth functions, and their associated Sobolev spaces -
that can be efficiently approximated with trigonometric polynomials.
We will hereafter refer to these inverse approximation theorems as
\emph{Bernstein-type} results.

\subsection{Sequence modelling as an approximation problem}
\label{subsec:formulation_seq}

Now, we introduce the problem of sequence approximation,
which can be regarded as a particular class of approximation problems
as introduced in~\cref{subsec:formulation_approx}.
The key difference with classical approximation theories is that the
input spaces $\set{X}$ and the output spaces $\set{Y}$
are now spaces of sequences, and may be infinite-dimensional.

We consider an input sequence indexed by a completely ordered index set $\set{T}$
\begin{equation}
    \seq{x}
    =
    \{
        x(t):
        t \in \set{T}
    \}.
\end{equation}
There are two main choices of the index set $\set{T}$.
For discrete sequences (e.g. sequences of word embeddings), $\set{T}$ is (a subset of) $\Z$.
For continuous sequences (e.g. measurements of a continuous-time control system),
$\set{T}$ is (a subset of) $\R$.
The input space is a collection $\set{X}$ of such sequences.
Correspondingly, the output space $\set{Y}$ is another collection of sequences.

Each input sequence $\seq{x} \in \set{X}$ corresponds to an output sequence
$\seq{y}$ with
\begin{equation}
    y(t) = H_t(\seq{x}),
    \qquad
    t\in\set{T}.
\end{equation}
That is, the sequence $\seq{H} = \{ H(t)\equiv H_t : t\in\set{T} \}$ is our target.
In this case, the target is in general an infinite-dimensional operator mapping $\set{X} \to \set{Y}$,
and for each $t$, $H_t$ is a functional on $\set{X}$.
We will hereafter refer to operators of this type as functional sequences.

Now, we seek to approximate $\seq{H}$ by candidates from a hypothesis space $\set{H}$.
The latter may be recurrent neural networks, convolutional networks or other types of models.
In each case, one first identifies appropriate target spaces $\set{C}$ for which $\set{H}$ is dense.
Then, one seeks Jackson-type and Bernstein-type results that characterize the types of sequence
relationships that can be efficiently approximated by each hypothesis space.

From the viewpoint of classical approximation theory,
one novel aspect of sequence approximation is that the input and output spaces are infinite-dimensional,
provided that the index set $\set{T}$ is infinite.
In fact, many interesting aspects of sequence modelling, such as those associated with memory,
precisely result from an unbounded index set, e.g. $\set{T} = \R$ or $\Z$.
We note that while sequence modelling is in effect an infinite-dimensional approximation problem,
it should be contrasted with generic operator learning problems
~\citep{kovachki2022.NeuralOperatorLearning,neufeld2022.ChaoticHedgingIterated,lu2021.LearningNonlinearOperators,chen1995.ApproximationCapabilityFunctions,benth2023.NeuralNetworksFrechet,cuchiero2022.UniversalApproximationTheorems,stinchcombe1999.NeuralNetworkApproximation}.
Here, a sequence is not a generic function but one with domain being a completely ordered index set.
Therefore, the sequences and their corresponding vector spaces contain temporal structure
that should be highlighted in the approximation results.
It is for this reason that we do not call $\seq{H}$ operators,
but functional sequences, to highlight the presence of its sequential structure.
We close this part with a final remark.
There are many applications where the output is not a sequence but rather just a finite dimensional vector.
Examples include sequence regression~\citep{xing2010.BriefSurveySequence}
and sentiment analysis~\citep{tang2015.DocumentModelingGated}.
The present formulation includes these cases by writing $y \equiv y(\infty) = H_{\infty}(\seq{x})$ as the input-output
functional relationship.

In next sections, we give a brief but structured overview of the approximation results for
sequence modelling, paying particular attention to the theoretical insights
and their consequences on practical architecture design.


%% file: rnn.tex
\section{Recurrent neural networks}
\label{sec:rnn}


Recurrent neural networks (RNN) are one of the earliest model
architectures proposed for modelling
sequential relationships~\citep{rumelhart1986.LearningRepresentationsBackpropagating}.
The key idea is the introduction of a hidden dynamical system
that captures the memory patterns in the sequences.
We begin by introducing the RNN architecture and its corresponding hypothesis space.

\subsection{Recurrent neural network hypothesis space}

We first consider modelling a sequential relationship on the
index set $\set{T}=\Z$.
Suppose at each time, the input sequence $x(t) \in \R^d$ is a vector.
Without much loss in generality, we can consider the output sequence
as a scalar sequence, i.e. $y(t) \in \R$.
For vector-valued output sequences, one may consider each output
dimension separately to deduce corresponding results.
The (one-layer) recurrent neural network parametrizes the relationship
between an input sequence $\seq{x}$ and an output sequence $\seq{y}$
as the following discrete dynamical system%
\footnote{
    There are notational variants in the literature,
    e.g. sometimes the index for the input is $t-1$ instead of $t$.
    Such minor variations do not affect approximation results.
}
\begin{equation}\label{eq:rnn_discrete_dynamics}
    \begin{aligned}
        h(t+1) &= \sigma(W h(t) + U x(t) + b), \\
        y(t) &= c^\top h(t),
    \end{aligned}
    \qquad
    t \in \Z.
\end{equation}
Here, $\seq{h}$ is a hidden state sequence, with each $h(t) \in \R^m$.
Thus, the trainable parameters are $W\in\R^{m\times m}$,
$U\in\R^{m\times d}$, $b\in\R^m$, and $c\in\R^m$.
Conventionally, we impose a zero initial condition on $\seq{h}$,
i.e. if the input sequence first becomes non-zero at a particular $t_0$
then $h(t_0) = 0$.
For theoretical treatments, we can also take $t_0 = -\infty$
to handle inputs of unbounded support.
The function $\sigma$ is an activation function, which is a scalar
function acting element-wise.
In typical RNNs, $\sigma$ is taken as the hyperbolic tangent function ($\tanh$),
but many other choices are possible.

Observe that \cref{eq:rnn_discrete_dynamics} defines a functional sequence
$\h{\seq{H}}$, with $y(t) = \h{H}_t(\seq x) = c^\top h(t)$,
and $h(t)$ satisfies the dynamics in \cref{eq:rnn_discrete_dynamics}.
Formally, we can write the RNN hypothesis space as
\begin{equation}\label{eq:Hrnn_discrete}
    \begin{aligned}
        \Hrnn
        &=
        \bigcup_{m \in \mathbb{N}_+}
        \Hrnn^m \\
        \Hrnn^m
        &=
        \left\{
            \begin{aligned}
            \seq{\h{H}}
            :
            &\h{H}_t(\seq x)
            =
            c^\top h(t),
            \seq{h} \text{ follows } \cref{eq:rnn_discrete_dynamics}
            \text{ with } \\
            &W\in \R^{m\times m},
            U\in \R^{m\times d},
            b\in \R^{m},
            c\in \R^{m}
            \end{aligned}
        \right\}
    \end{aligned}
\end{equation}
The approximation budget here is $m$, which is the width of the RNN,
or the dimension of the hidden state sequence $\seq{h}$.
Approximation theory of RNN investigates the ability of $\Hrnn$ and $\{\Hrnn^m\}$
to approximate appropriate target functional sequences.

It is often convenient to consider a continuous variant of the RNN, i.e. $\set{T} = \R$.
In this case, the RNN hidden state equation is now continuous in $t$, and it can be viewed
as a time index.
The only change is that we replace the difference equation \cref{eq:rnn_discrete_dynamics}
by the differential equation
\begin{equation}\label{eq:rnn_cts_dynamics}
    \begin{aligned}
        \dot{h}(t) &= \sigma(W h(t) + U x(t) + b), \\
        y(t) &= c^\top h(t),
    \end{aligned}
    \qquad
    t \in \R.
\end{equation}
Besides theoretical advantages, some practical applications (e.g. irregularly-sampled time series)
require a continuous-index model.
The corresponding hypothesis space is analogous to \cref{eq:Hrnn_discrete} with \cref{eq:rnn_cts_dynamics}
in place of \cref{eq:rnn_discrete_dynamics}.

A remark is in order on the choice of time-index for sequence approximation.
Generally, $\set{T}$ can be discrete or continuous, and bounded or unbounded, leading to four different settings.
In addition, in each setting there is a choice of the norm that measures the approximation error.
In the simplest case where $\set{T}$ is bounded and discrete, the approximation problem is finite-dimensional.
Beyond this setting, the choice of norm generally matters.
For density-type results, the choice of discrete vs continuous $\set{T}$ is usually not important,
since they can be bridged by a discretization argument in one way and taking limits in the other.
The distinction between bounded and unbounded $\set{T}$ is however significant, and the latter
generally requires more stringent conditions and is also more important for analyzing
memory behavior that occurs at asymptotic regimes of $\set{T}$.
On the other hand, for Jackson/Bernstein-type theorems, there is a difference between
discrete and continuous $\set{T}$.
Typically, approximation error estimates for a discrete $\set{T}$ are grid-dependent,
and do not readily translate to a uniform error estimate over all discrete grid partitions.
In this sense, uniform-in-$t$ estimates for the continuous case are stronger results,
as they imply approximation rates for any grid using a discretization argument,
given some regularity conditions on the sequences to allow one to estimate the discretization error.

\subsection{Density-type results}
\label{sec:rnn_density}


As with most machine learning models, density-type results are the most basic and thus prevalent.
Such results are minimal guarantees for the general applicability of a machine learning model.
At the same time, the most theoretically interesting part about these results is
the identification of appropriate target spaces $\set{C}$ in which a particular $\set{H}$ is dense.

\paragraph{Hidden dynamic functionals.}
By observing the RNN structure, it is natural to consider target functionals that are themselves
defined via observations of a hidden dynamical system that has a compatible structure.
For instance, in continuous time index case one can consider
\begin{equation}\label{eq:fg_dynamics}
    \seq x \mapsto \seq{H}(\seq{x}) = \seq{y}
    \quad
    \text{with}
    \quad
    \begin{aligned}
        \dot{h}(t) &= f(h(t), x(t)), & h(t) &\in \R^{n},\\
        y(t) &= g(h(t)), & h(-\infty) &= 0,
    \end{aligned}
\end{equation}
where $f:\R^n \times \R^d \to \R^n$ and $g:\R^n \to \R$.
We may assume that $f$ is Lipschitz and $g$ is continuous so that
$\seq{H}$ is well-behaved.
The function $g$ is called a readout map.
Since the functions $f,g$ parameterize a functional
sequence via a hidden dynamical system,
we call them \emph{\fg},
or $\Cfg$ for short.
The discrete or bounded index cases are defined similarly.
In the non-linear dynamics literature, \cref{eq:fg_dynamics}
is often called a non-linear time-invariant system
and the corresponding functional sequence $\seq{H}$ is
referred to as a time-invariant filter.
The term time-invariant (strictly, equivariant) highlights that $\seq{H}$ commutes with time-shifts.
To see this, denote by $\seq{S}_\tau$ the shift operator $\seq{S}_\tau(\seq{x})(t) = x(t-\tau)$,
then $\seq{H}$ satisfies $\seq{H} \circ \seq{S}_\tau = \seq{S}_\tau \circ \seq{H}$.
However, in this survey we refrain from calling them time-invariant filters,
because there may exist functional sequences that commute with time shifts,
but are not readily written in the form \cref{eq:fg_dynamics},
e.g. the shift functional sequence $H_t(\seq{x}) = \seq{S}_\tau(\seq{x})(t) = x({t-\tau})$.

\rev{
Density-type results for $\Cfg$ are also called
\emph{universal simulation},
since it requires the approximate simulation of a dynamics driven by $f$
and a readout map defined by $g$ by a RNN.
Earlier results on {\fg} focus on a bounded index set
(see the survey of~\citet{sontag1992.NeuralNetsSystems} and references therein,
and also~\citet{chow2000.ModelingContinuousTime,li2005.ApproximationDynamicalTimevariant}).
In these works, the main technique is to appeal to the universal approximation
theory of feed-forward networks (e.g. \citet{cybenko1989.ApproximationSuperpositionsSigmoidal}).
The simple observation is that the right hand side of RNNs
are feature maps of a fully connected network.
Thus, by increasing $m$ one can construct an approximation of $f$ as
\begin{equation}\label{eq:rnn_rhs}
    (h, x) \mapsto f(h, x)  \approx (h_1, x) \mapsto \sigma(W(h_1, h_2)^\top + Ux + b),
\end{equation}
where $h_1 \in \R^n$ and $h_2 \in \R^{m-n}$.
The readout map $g$ can be handled likewise.
A similar approach is developed in~\citet{schafer2006.RecurrentNeuralNetworks}
in the discrete-time setting,
and~\citet{funahashi1993.ApproximationDynamicalSystems} for simulating
dynamics without inputs.
Since the results concern a compact time interval,
to approximate dynamics it is enough to approximate $f$.
This is in general not true for the unbounded case,
as the approximation error can be magnified by the dynamics.}

To handle unbounded $\set{T}$ (e.g. $\set{T}=\R$),
one strategy is to introduce
some decay properties to the targets.
One such property is the \emph{fading memory property} (FMP)
\citep{boyd1985.FadingMemoryProblema}.
Let $\seq{x_1}, \seq{x_2}$ be bounded sequences indexed by $\R$,
and let $\seq{H}$ be a sequence of causal, shift-equivariant
(also called time-homogeneous) functionals.
Here, causal means $H_t(\seq{x}) = H_t(\seq{x}_{(-\infty,t]})$ for all $t$.
We say that $\seq{H}$ has the FMP if
there is a monotonically decreasing function $\seq{w}:\R_+ \to (0, 1]$
such that for any $\epsilon > 0$ there exists $\delta > 0$
with
\begin{equation}\label{eq:fmp}
    |H_t(\seq{x_1}) - H_t(\seq{x_2})| < \epsilon
    \text{ whenever }
    \sup_{s \in (-\infty, t]} |x_1(s) - x_2(s)| w(t-s)
    <
    \delta.
\end{equation}
Intuitively, this says that two inputs sequences
that differ more and more in their history ($t\to-\infty$)
still produce similar outputs at the present.
This is in effect requiring the memory of $\seq{H}$ to decay.
Note that due to time-equivariance, it is enough to check
this for just one $t$, say $t=0$.
Then, we can define a weighted norm on the space of
semi-infinite sequences on $(-\infty, 0]$ by
\begin{equation}
    \| \seq{x} \|_{\seq w}
    =
    \sup_{s\in (-\infty, 0]}
    |
        x(s) w(-s)
    |.
\end{equation}
Consequently, the FMP (\cref{eq:fmp}) is simply a continuity requirement
of $H_0$ with respect to $\normw{\cdot}$.
We denote by $\Cfmp$ the set of causal, shift-equivariant
functional sequences satisfying the FMP.
The FMP allows one to prove density on unbounded $\set{T}$,
e.g. in~\citet{grigoryeva2018.EchoStateNetworks}
and~\citet{gonon2021.FadingMemoryEcho}.
Indeed, the FMP property allows one to approximate $\seq{H} \in \Cfmp$
by a truncated version on a bounded interval.
Then, approximation results can be deduced using methodologies
for the bounded case.
Note that the FMP is defined for general functional sequences,
and is not limited to the form of \fg.
Thus, this idea can also be used to prove density
for general functionals on unbounded index sets.

\rev{
    In the specific setting of \fg, another technique for handling unbounded index sets was
    proposed in~\citet{hanson2020.UniversalSimulationStable}.
    Here, the authors consider dynamics driven by
    $f$ that satisfy a property called
    ``uniformly asymptotically incrementally stable''.
    This roughly says that the flow maps of $\dot{h}=f(h, x)$ are uniformly
    continuous, uniformly in $\seq{x}$, and that $h(t)$ is independent
    of initial condition at large $t$.
    One can understand this as again a memory decay condition,
    as any initial condition on $h$ is forgotten in the large time limit.
    This allows one to localize the approximation of $f$ and $g$
    to a compact set, which then allows one to appeal to standard
    approximation results from the feed-forward networks.
}

\paragraph{General functional sequences.}

Now, we turn to more general functional sequences.
Since the RNN architecture (\cref{eq:Hrnn_discrete}) is causal
and shift-equivariant, we should restrict our attention
to target spaces satisfying the same properties.
However, we no longer assume that
these functional sequences admit a representation in the form of
\cref{eq:fg_dynamics}.
For density-type results, this distinction is not important.
This is because it is known that $\Cfg$ is dense in $\Cfmp$ in the norm
$\| \seq{H} \| = \sup_{t\in\R,\seq{x}\in \set{K}} |H_t(\seq{x})|$,
where $\set{K}$ is a bounded equicontinuous set in $C(\R)$%
~\citep[Thm.~2]{boyd1984.AnalyticalFoundationsVolterra};
see also~\citet{grigoryeva2019.DifferentiableReservoirComputing}.
The idea relies on approximation of FMP functionals by a
Volterra series~\citep{volterra1930.TheoryFunctionalsIntegral}.
The density can also be established without appealing to
the Volterra series~\citep[Thm.~8]{grigoryeva2018.EchoStateNetworks}.
Therefore, density-type results on $\Cfg$ can be passed onto $\Cfmp$,
provided the norms are compatible.
In the RNN case, this program is carried out in%
~\citet{grigoryeva2018.UniversalDiscretetimeReservoir,grigoryeva2018.EchoStateNetworks}.
However, we will see later that for Jackson-type results,
the choice of target spaces is important:
the rate of approximations generally depends on such choices.

It is also possible to construct a RNN approximation in $\Cfmp$ directly,
without the need to use $\Cfg$ as an intermediate.
For example, in~\citet{gonon2021.FadingMemoryEcho} the authors
first use the FMP to reduce the approximation problem to one over
a finite, bounded index set, and then appeal to the density of
fully connected neural network to obtain approximation.
It remains then to construct a (large) RNN to represent the
fully connected network.
A similar result for stochastic inputs is proved in
\citet{gonon2018.ReservoirComputingUniversality}.

Many of the aforementioned density-results stem from the
\emph{reservoir computing} literature,
where researchers are interested in studying systems
such as the RNN, but with the internal weights
($W,U,b$ in \cref{eq:rnn_cts_dynamics})
being random variables.
This random version of the RNN is called an echo-state network (ESN).
From the machine learning viewpoint,
one can understand ESNs as an analogue of random feature models
corresponding to RNNs.
These models have the nice property that the hypothesis space is linear
and training these networks is a convex problem,
since only $c$ needs to be trained.
Previously mentioned results show existence of $(W,U,b)$ and $c$
to approximate each $\seq{H}$, but do not address the approximation
of classes of $\seq{H}$ by choosing only $c$ and using a
common random realization of $(W,U,b)$.
The latter approximation problem is studied in%
~\citet{gonon2021.ApproximationBoundsRandom},
where a density result with some explicit error estimates
is obtained.
Here, the primary idea is to constrain target functionals
to a subset of $\Cfmp$ whose Fourier transform has finite third moment.
\rev{
    This builds on the idea of \citet{barron1992.NeuralNetApproximation,barron1993.UniversalApproximationBounds}
    where functions of this type (but with finite first and second moments)
    were shown to be approximated by feed-forward neural networks without suffering the curse of dimensionality.
    This is to be contrasted with a related line of work%
    ~\citep{e2019.PrioriEstimatesPopulation,e2020.MathematicalUnderstandingNeural,wojtowytsch2020.BanachSpacesAssociated},
    which introduces a probabilistic definition of Barron-type functions via an expectation
    in place of a moment condition on its Fourier transform.
    In both cases, it is known that such functions can be approximated by randomly sampling
    neural network weights according to a distribution to achieve approximation.
    This is used in~\citet{gonon2021.ApproximationBoundsRandom} to prove
    density for ESNs with random weights.
    We note that in general Barron function approximations,
    the random weight distributions depend on the target functions to be approximated,
    whereas in~\citet{gonon2021.ApproximationBoundsRandom} the distribution
    of the reservoir weights is fixed as uniform.
    This comes with the cost of stronger regularity conditions,
    as we will discuss later.
}

\subsection{Jackson-type results}
\label{subsec:rnn_jackson}


\rev{
    Compared to density-type results,
    there are fewer Jackson-type results for RNNs.
    In the aforementioned work of~\citet{gonon2021.ApproximationBoundsRandom},
    a quantitative error estimate can be obtained by a time-truncation argument
    in the discrete time index setting.
    Let $\seq{H}|_T$ denote the restriction of $\seq{H}$
    to sequences of length $T+1$.
    Then, for each $t$ we can identify $H|_T(t)$ with a function
    $H_{T,t} : \R^{d\times(T+1)} \rightarrow \R$.
    If one imposes additional regularity conditions by requiring
    $H_{T,t} \in \mathcal{W}^{k,2}$ for each $t$, then
    one can deduce an error estimate of the form
    \begin{equation}
        \inf_{\seq{\h{H}} \in \Hesn^m}
        \E
        \left[
            \| \seq{H} - \seq{\h{H}} \|^2
        \right]^{1/2}
            \leq
            c_1
            \frac{\| \seq{H}|_T \|_{W^{k,2}}}{m^{1/\alpha}}
            +
            c_2
            \sum_{i=T+1}^{\infty} w(-i),
    \end{equation}
    where $\alpha > 2$ and $w$ is the weighting function
    used in the definition of $\Cfmp$.
    In particular, if we consider approximation on a
    bounded index set the last term vanishes,
    and we obtain an approximate Monte-Carlo rate $1/\sqrt{m}$.
    However, a caveat is that the smoothness requirement $k$
    for this estimate to hold increases linearly with $dT$,
    i.e. it becomes increasingly stringent on larger time intervals
    or input dimensions.
    In other words, this estimate is more useful for bounded
    index sets and low input dimensions.

    In the setting of {\fg}, a similar estimate is proved in~\citet{hanson2020.UniversalSimulationStable}
    for unbounded index sets.
    The key assumption of uniformly asymptotically incrementally stable dynamics
    (c.f. the discussion in \cref{sec:rnn_density})
    is combined with the additional assumption that $f,g$ are Barron-type functions.
    Then, one can obtain a Monte-Carlo error rate that decays as $1/\sqrt{m}$.
    The argument is a combination of the localization argument outlined
    previously for the density result, and the application of the results of
    ~\citet{barron1992.NeuralNetApproximation,barron1993.UniversalApproximationBounds}
    on the localized compact domain.
}

A general property of these results is the reliance on time truncation,
thus the rate estimates do not explicitly account for the behavior on large
time intervals.
Jackson-type error estimates that directly operates on unbounded time domains
are proved in the linear RNN case
($\sigma(z) = z$ and $b=0$ in \labelcref{eq:rnn_cts_dynamics})%
~\citep{li2021.CurseMemoryRecurrent,li2022.ApproximationOptimizationTheorya}.
Let us call these hypothesis spaces $\Hrnnl$ and $\{\Hrnnl^m\}$.
Observe that each $\seq{\h{H}} \in \Hrnnl^m$ has the form
\begin{equation}\label{eq:hlrnn}
    \h{H}_t(\seq{x})
    =
    \int_{0}^{\infty}
    c^\top
    e^{Ws}
    U x(t-s)
    ds,
    \qquad
    c \in \R^{m},
    W \in \R^{m\times m},
    U \in \R^{m\times d}.
\end{equation}
Here, the input space considered is $\set{X} = C_0(\R, \R^d)$,
the space of continuous vector-valued sequences
vanishing at infinity.
We will also assume that $W$ is Hurwitz
(i.e. it is non-singular with eigenvalues having negative real parts),
so that the dynamics is stable.
In this case, one can check that each $\seq{\h H}$ is linear,
continuous in the uniform norm and shift-equivariant (time-homogeneous).
In addition, it is regular in the sense that if $\seq{x}_n(t) \to 0$
for almost every $t$ then $\seq{H}(\seq{x}_n) \to 0$.
It turns out that that these conditions are sufficient conditions
for functionals in $\set{C}$ to be uniformly approximated by
linear RNNs~\citep{li2022.ApproximationOptimizationTheorya}.
The idea is straightforward:
one first shows that any linear functional sequence $\seq{H}$
satisfying these conditions admits a common Riesz representation
\begin{equation}\label{eq:commonriesz}
    H_t(\seq{x})
    =
    \int_{-\infty}^{t}
    \rho(t-s)^\top x(s) ds
    =
    \int_{0}^{\infty}
    \rho(s)^\top x(t-s) ds.
\end{equation}
In other words, $\seq{H}$ and $\seq{\rho} \in L^1$ can be identified.
Note that the application of Riesz representation is valid since
$C_0(\R,\R^d)$ is taken as the input sequence space.
In broader settings, e.g. $C(\R,\R^d)$ where input sequences
need not decay at infinity,
more assumptions is required for the existence of this representation.
For example, \citet[Thm.~5]{boyd1985.FadingMemoryProblema}
shows that if $\set{X} = C(\R,\R)$,
$\seq{H}$ admits the form \eqref{eq:commonriesz}
if and only if $\seq{H}$ has fading memory,
in addition to the aforementioned assumptions.
Now, comparing \cref{eq:hlrnn} and \cref{eq:commonriesz},
linear RNN approximation of these functionals boils down to
\begin{equation}
    |H_t(\seq{x}) - \h{H}_t(\seq{x})|
    \leq
    \| \seq{x} \|_{L^\infty}
    \| \seq{\rho} - \seq{\h{\rho}} \|_{L^1},
\end{equation}
where $\h{\rho}(s) = [c^\top e^{Ws} U]^\top$.
Therefore, we may deduce approximation properties of targets
by linear RNNs by approximation of functions in $L^1$
by exponential sums of the form $[c^\top e^{Ws} U]^\top$.
The density of such exponential sums can be derived using
the M\"{u}ntz–Sz\'{a}sz theorem~\citep{lorentz2005.ApproximationFunctions}.

Similarly, ~\citet{li2022.ApproximationOptimizationTheorya}
further use this idea to prove a Jackson-type result for the error estimate.
Here enters the crucial property of memory decay.
There exists a vast literature on possible notions of memory decay
for functional sequences, see e.g.~\cite{boyd1985.FadingMemoryProblema,gonon2020.ReservoirComputingUniversality} and references therein.
In the linear case, the following simple definition suffices.
Let $\seq{e}_i = e_i \mathbf{1}_{t\geq 0}$,
$i=1,\dots,d$ with $e_i$ the unit vector in the $i$-th axis direction.
We consider targets $\seq{H}$ such that there exist
$\alpha \in \Z_+$, $\beta > 0$ such that
\begin{equation}\label{eq:efmp}
    e^{\beta t} H^{(r)}_t(\seq{e_i})
    =
    o(1),
    \qquad
    t \to \infty,
    \qquad
    i = 1, \dots, d,
    \quad
    1 \leq r \leq \alpha+1.
\end{equation}
Intuitively, these functionals forget input history
at a rate of at least $e^{-\beta t}$.
Thus, we may also understand them possessing
an \emph{exponentially} decaying memory.
The main result in~\citet{li2021.CurseMemoryRecurrent}
is a Jackson-type error estimate
\begin{equation}\label{eq:jackson_rnn}
    \inf_{\seq{\h H} \in \Hrnnl^m}
    \| \seq{H} - \seq{\h H} \|
    \leq
    \frac{c_\alpha d \gamma}{\beta m^\alpha},
    \qquad
    \gamma =
    \sup_{t\geq 0}
    \max_{i=1,\dots,d}
    \max_{r=1,\dots,\alpha+1}
    \frac{|e^{\beta t} H^{(r)}_t(\seq{e}_i)|}{\beta^r},
\end{equation}
where $\|\seq{H}\| = \sup_{t} \sup_{\|\seq{x}\|_{L^\infty} \leq 1} |H_t(\seq{x})|$.
Comparing with \cref{eq:jackson_trig}, we see that
one obtains a similar rate characterized by the smoothness parameter
$\alpha$.
The new phenomena is the assumption of exponential decaying memory
in \cref{eq:efmp}.
The key insight here is as follows.
If we assume, in addition to the usual smoothness requirements,
that the memory of targets decay like an exponential
(\cref{eq:efmp}), then we can efficiently approximate them using
linear RNNs.

We remark here that this result demonstrates the importance of
considering more general functional sequences than $\Cfg$
in establishing Jackson-type results.
Assume instead that one considers {\fg} with both $f,g$
as linear functions, i.e.
\begin{equation}
    f(h, x) = W_* h + B_* x,
    \quad
    g(h) = c_*^\top h,
    \qquad
    W_*\in\R^{n\times n},
    U_*\in\R^{n\times d},
    c_*\in\R^{n}.
\end{equation}
Then, the rate estimate becomes trivial:
If $m \geq n$, then the approximation error is 0
and we have perfect representation.
However, in practice it is generally not possible
to know the precise mechanism for the generation
of the sequence data,
and a theory should handle general functional sequences.
From the Riesz representation \labelcref{eq:commonriesz}
of general linear, causal and shift-equivariant target functional sequences,
$\seq \rho$ can be any $L^1$ function, and
may not in the form of an exponential sum.
In this case, the approximation rate estimate
becomes non-trivial.

We close the discussion by discussing the so-called
\emph{curse of memory} phenomenon identified in the above analysis
observed in~\citet{li2021.CurseMemoryRecurrent}.
The density type results, including the linear RNN case,
do not require the targets to have an exponentially decaying memory in the sense of \cref{eq:efmp}.
However, the rate estimate in \cref{eq:jackson_rnn} does have this requirement.
The natural question is therefore, what if one has a slower memory decay rate?
For example, we may replace \cref{eq:efmp} by
\begin{equation}
    H^{(r)}_t(\seq{e_i})
    \sim
    e^{-\beta t}
    \quad
    \longrightarrow
    \quad
    H^{(r)}_t(\seq{e_i})
    \sim
    t^{-(r+\omega)}
    \quad
    (\omega > 0).
\end{equation}
Then, a truncation argument in~\citet{li2022.ApproximationOptimizationTheorya}
shows that to obtain an approximation error of $\epsilon$,
a size of the RNN may need to grow exponentially,
as $m \sim \epsilon^{-1/\omega}$.
While this is not a lower bound for the optimal approximation error,
it suggests that in sequence approximation,
one may observe a very similar issue with approximating ordinary functions
in high dimensions.
There, it is known that the approximation budget required to
achieve a prescribed approximation error grows
like an exponential function of the dimension of the function domain.
This is known as the \emph{curse of dimensionality}.
Here, the results suggests that in sequence approximation problems using RNNs,
there lies a \emph{curse of memory}.
In particular, it affirms the empirical observations that RNNs
usually perform well when memory in the system is small,
but suffer in its performance for approximating long-term memory%
~\citep{bengio1994.LearningLongtermDependencies}.
The result in \cref{eq:jackson_rnn} confirms the first part of
the observation.
The second part can be further demonstrated by optimization analysis
\citep{li2022.ApproximationOptimizationTheorya}
and also a Bernstein-type result, as we discuss next.

\subsection{Bernstein-type results}


Recall that Bernstein-type results deduce properties
of targets assuming that they can be efficiently approximated
by a hypothesis space.
Known Bernstein-type results for RNNs are currently
limited to linear functional sequences.
With the same set-up as the Jackson-type theorem,
\citet{li2022.ApproximationOptimizationTheorya}
proves a Bernstein-type result, which we now describe.

Let us assume that we have a target functional sequence $\seq{H}$
such that it (and its derivatives in time) can be uniformly
approximated by a sequence of linear RNNs.
That is, we assume that there is a sequence
$\seq{\h H}_m \in \Hrnnl^m$ such that
$\| \seq{H} - \seq{\h H}_m \| \to 0$ and that
\begin{equation}
    \sup_{t\geq 0}
    |
        H_t^{(k)}(\seq{e}_i)
        -
        {\h{H}}_{m, t}^{(k)}(\seq{e}_i)
    |
    \to 0,
    \qquad
    k = 1,\dots,\alpha+1.
\end{equation}
Then, under additional technical conditions,
there must exist a $\beta > 0$
such that
\begin{equation}\label{eq:berstein_rnn}
    e^{\beta t} H^{(r)}_t(\seq{e_i})
    =
    o(1),
    \qquad
    t \to \infty,
    \qquad
    i = 1, \dots, d,
    \quad
    1 \leq r \leq \alpha + 1.
\end{equation}
In other words, a target can be effectively approximated
by linear RNNs only if it has exponentially decaying memory.
This is in a sense a partial converse to the Jackson-type result
in \cref{eq:jackson_rnn}.
Together, it shows that, at least in the linear setting,
effective RNN approximation occurs if and only if
the target functional sequence has an exponentially decaying memory pattern.
Bernstein-type results can assist in designing architectures
for sequence modelling:
if a model aims to model a sequential relationship whose memory
pattern does not decay like an exponential,
then it is necessary to go beyond the RNN setting due to the limitations
posed by the inverse approximation result.
\rev{
    At the end of \cref{sec:others_cnn},
    we discuss an example given in~\citet{jiang2021.ApproximationTheoryConvolutional}
    where the target functional sequence does not have an exponentially decaying memory,
    and alternative architectures such as dilated convolutions are shown to be
    more effective than RNNs.
}

We end the section on RNNs approximation by discussing some of its variants.
In the practical literature, a number of generalizations
of the simple RNN hypothesis space (\cref{eq:Hrnn_discrete})
have been proposed.
Examples include the long-short term memory (LSTM) network~\citep{hochreiter1997.LongShortTermMemory}
and gated recurrent units (GRU)~\citep{cho2014.PropertiesNeuralMachine}.
Density-type results for these networks can be directly deduced
since they often include the classical RNN as a special case
by a proper choice of its trainable parameters.
In some cases
(%
    e.g. normalized RNNs in~\citet{schafer2006.RecurrentNeuralNetworks},
    and deep variants with fixed width in~\citet{song2022.MinimalWidthUniversal}%
),
additional analysis is required to establish density.
However, rate estimates of Jackson-type or inverse theorems
of Bernstein-type (different from classical RNNs)
are generally not known for these more complex structures,
and is an interesting direction of future work.

%% file: others.tex
\section{Other architectures}
\label{sec:others}

Let us now expand our discussion to models beyond
the RNN model family.
Many of these architectures are proposed or popularized
in fairly recent years.
A partial but important motivations for developing these
alternative model architectures is precisely the limitations
with respect to memory we have discussed in~\cref{sec:rnn}.
Very often in practical applications,
we want to model sequence relationships having long
and irregular memory patterns.
For example, in machine translation tasks,
an output word at the end of the sentence in one language
may depend on the very first word in the corresponding sentence
in another language.
Moreover, the number of words in the original and translated
sentences are often not the same.
For these reasons, a variety of alternative models to the RNN
have been proposed.
Each of them are competitive in different domains of application.
The subsequent discussions will highlight a number of such examples.

However, to concretely understand the gains of using alternative
architectures to RNN, it is necessary to develop some
theoretical understanding of their comparison.
For example, can an alternative architecture such as
a convolutional-based architecture overcome the
curse of memory related to RNNs?
This often requires the developments of
Jackson-type estimates in similar approximation settings,
which tells us precisely which functional sequences are easy to approximate
under a particular hypothesis space corresponding to
a model architecture of interest.

\subsection{Convolution-based architectures}
\label{sec:others_cnn}


We begin with results for convolutional-based architectures.
While convolutional neural networks (CNN) were originally developed
for computer vision applications
\citep{krizhevsky2017.ImageNetClassificationDeep},
temporal versions of the CNNs have been shown to be
effective in many sequence modelling tasks~\citep{bai2018.EmpiricalEvaluationGeneric}.
Since convolution operations are easier to describe
using a discrete index set, we shall assume throughout
this subsection that $\set{T} = \Z$.

The basic building block of temporal CNNs is
the causal dilated convolution operation
\begin{equation}\label{eq:dilated_conv}
    (\seq{u} \Conv{}_{l} \, \seq{v})(t)
    =
    \sum_{s \geq 0} u(s)^\top v(t-ls),
    \qquad
    l \in \Z_+.
\end{equation}
Note that the summation is taken over $s\geq 0$
to ensure causality,
meaning that the outcome at time $t$ depends only on the past information.
When $l=1$, this is the usual convolution.
Dilations $l\geq 2$ result in larger receptive fields with the same
number of parameters, and are hence useful in processing long sequences.
For example, successful temporal CNN architectures,
including the WaveNet~\citep{oord2016.WaveNetGenerativeModela}
and the TCN~\citep{lea2017.TemporalConvolutionalNetworks},
contain stacks of dilated convolutions with increasing dilation rates.

We can write a general dilated temporal CNN model with
$K$ layers and $M$ channels at each layer as
\rev{\begin{equation}\label{eq:CNNdynamics}
		\begin{aligned}
		\seq h_{0,i}
        &=
        \seq x_i,\\
		\seq h_{k+1,i}
        &=
        \sigma \left(\sum_{j=1}^{M_k} {\seq w}_{kji}
        \Conv{}_{d_k} \seq h_{k,j} \right),
        \quad i=1, \dots, M_{k+1},
        \quad k=0, \dots, K-1\\
		\seq {\h y} &= \seq h_{K,1},
		\end{aligned}
\end{equation}
where $M_0=d$ is the input dimension,
$M_K=1$ is the output dimension.
$M_k=M$ is the number of channels at layer k for $k=1, \dots, K-1$.
}
Here, $\seq x_{i}$ is the scalar sequence
corresponding to the $i^\text{th}$ element
of the vector sequence $\seq x$,
and ${\seq w}_{kji}$ is the convolutional filter at layer $k$,
mapping from channel $j$ at layer $k$ to
channel $i$ at layer $k+1$.
A common choice for the dilation rate in applications is $d_k = 2^K$,
so we adopt this choice for the subsequent exposition.
Furthermore, for establishing approximation results
it is sufficient to assume that the support of each filter
$\seq{w}_{kji}$ is 2, since convolutional filters
of large sizes include this case.
This gives rise to the temporal CNN hypothesis space
\begin{equation}
    \Hcnn =
    \bigcup_{K, M}
    \Hcnn^{(K,M)}
    =
    \bigcup_{K, M}
    \Big\{
        \seq{x} \mapsto \seq{\h{y}}
        \text{ in \cref{eq:CNNdynamics}}
    \Big\}.
\end{equation}

Density-type results have been studied for general CNNs
mostly for two-dimensional image applications,
and some of them can be adapted to the one-dimensional,
causal case here.
For brevity, we will not give an exhaustive list of this literature.
We mention however that most existing results are not directly applicable
to the sequence modelling case due to the shift-equivariant requirement.
For example, the works of
\citet{
    oono2019.ApproximationNonparametricEstimation,
    zhou2020.UniversalityDeepConvolutional,
    okumoto2022.LearnabilityConvolutionalNeural}
consider approximating general functions,
and shift-equivariance is violated at the boundaries.
Density results for fully convolutional cases
\citep{
    li2022.DeepNeuralNetwork,
    lin2022.UniversalApproximationProperty,
    petersen2020.EquivalenceApproximationConvolutional,
    yarotsky2022.UniversalApproximationsInvariant}
are more relevant for the present application.
Nevertheless, due to the nature of image data having finite supports,
none of these results consider an unbounded index set.
However, for sequence approximation,
the problem of memory should be studied precisely on unbounded index sets.
If we assume some form of memory decay such as the FMP,
then a truncation argument can be used to show that the temporal
CNN hypothesis space is dense in sequence spaces (e.g. $\ell^p$),
as a corollary of these results.

For Jackson-type theorems, the current understanding is again
limited to the simple but interesting case of linear temporal CNNs,
i.e. $\sigma(z) = z$.
This gives the linear temporal CNN hypothesis space
\begin{equation}\label{eq:linearhcnn}
    \begin{aligned}
        \Hcnnl = \bigcup_{K, M}
        \Hcnnl^{(K,M)} = &\Big\{\seq{\h{H}}:
        \h{H}_t(\bm x) = \sum_{s=0}^\infty \h{\rho}(s)^\top x(t-s)
        \Big\},
    \end{aligned}
\end{equation}
where $\seq{\h{\rho}}$ is a finitely-supported
sequence determined by the filters $\{\seq{w}_{kji}\}$:
\begin{equation}\label{eq:cnn_representation}
    \seq{\h \rho}_i
    =
    \sum_{i_1, \dots, i_{K-1}}
    \seq{w}_{K-1, i_{K-1}, 1}
    \Conv{}_{2^{K-1}}
    \seq{w}_{K-2, i_{K-2}, i_{K-1}}
    \Conv{}_{2^{K-2}}
    \dots
    \Conv{}_{2}
    \seq{w}_{0, i, i_1}.
\end{equation}
Observe the striking similarity of~\labelcref{eq:linearhcnn}
and linear RNN hypothesis space~\labelcref{eq:hlrnn}.
The key difference is that in the RNN case,
the sequence $\seq{\h \rho}$ is an exponential sum with infinite support,
whereas in the case of CNNs it is a sum of repeated
dilated convolutions resulting in a finite support.
This in turn leads to, as investigated in
\citet{jiang2021.ApproximationTheoryConvolutional},
vastly different Jackson-type estimates.
In particular, one can identify different approximation
spaces that suggests how RNN and CNN approximation
differ when modelling sequence relationships.

Concretely,~\citet{jiang2021.ApproximationTheoryConvolutional}
proved the following Jackson-type estimate for linear,
causal and shift-equivariant
functional sequences $\seq{H}$:
\begin{equation}\label{eq:jackson_cnn}
    \inf_{\seq{\h{H}} \in \Hcnnl^{(K,M)}}
    \| \seq{H} - \seq{\h{H}} \|
    \leq
    G(KM^{\frac 1 K}-K)C_1(\seq{H})d  + C_2(\seq H, K).
\end{equation}
Recall that $M$ is the number of convolution filters at each layer
and $K$ is the number of layers.
Together, $(M,K)$ control the complexity of the CNN hypothesis space.
The function $G : \R \to \R$ is a non-increasing function tending to 0,
to be explained later.

Let us now clarify the form of $C_1,C_2$.
Let $\seq\rho$ be the Riesz representation of $\seq H$,
\begin{equation}
    H_t(\seq x) = \sum_{s\geq 0} \rho(s) x(t-s).
\end{equation}
Then, $C_2(\seq{H}, K)=\| \seq\rho_{[2^K, \infty) }\|_{\ell^2}$
is determined by the rate of decay of the memory
of the target functional sequence.
In particular, $C_2$ decays at least exponentially in
the depth of the neural network ($K$),
even if the target does not possess memory decay.
The term $C_1(\seq{H)}$ is a complexity measure of the target functional sequence,
determined by the effective rank of $\seq{H}$ after a tensorisation transformation.
Let us motivate its definition by an example.
Set $d=1$ and suppose the goal is to model a target functional sequence
\begin{equation}
    H_t(\seq{x})
    =
    r_0 x(t)
    +
    r_1 x(t-1)
    +
    r_2 x(t-2)
    +
    r_3 x(t-3),
    \qquad
    r_s \in \R.
\end{equation}
In this case, the Riesz representation for $\seq{H}$ has support 4, i.e.
$
    \seq{\rho}
    =
    (r_0, r_1, r_2, r_3)
$.
\rev{
A temporal CNN approximates $\seq{\rho}$ via product-sums in
the form of~\labelcref{eq:cnn_representation}.
Let us take $K=2$ and $M=1$.
Then, notice that we are seeking the approximation of
\begin{equation}
    \seq{\rho}
    =
    (r_0, r_1, r_2, r_3)
    \qquad
    \text{by}
    \qquad
    \seq{\h \rho}
    =
    (w_{0,0}, w_{0,1})
    \Conv{}_2
    (w_{1,0}, w_{1,1}),
\end{equation}
which we can rewrite in matrix form as the approximation of
\begin{equation}
    \tens{\seq{\rho}}
    =
    \begin{pmatrix}
        r_0 & r_1\\
        r_2 & r_3
    \end{pmatrix}
    \qquad
    \text{by}
    \qquad
    \tens{\seq{\h \rho}}
    =
    \begin{pmatrix}
        w_{0,0} \\
        w_{0,1}
    \end{pmatrix}
    \begin{pmatrix}
        w_{1,0} & w_{1,1}
    \end{pmatrix}.
\end{equation}
Then, the approximation error becomes clear.
If $\tens{\seq{\rho}}$ is rank 1, then it can be represented
exactly by the 2-layer CNN with channel size 1.
Otherwise, there will be an approximation error,
and the optimal approximation error is the second singular
value of $\tens{\seq\rho}$ as a consequence of the Eckart-Young theorem.

This argument can be generalized to any $K$ and $M$.
For $K \geq 3$ the reshaping operation $\tens{\cdot}$
acting on a length $2^K$ sequence produces an order-$K$ tensor of size $2$
in each dimension,
\begin{equation}
    \tens{\seq\rho_{[0,2^K]}}_{i_1,\dots,i_K}
    =
    \rho_{[0,2^K]}
    \left(
        \sum_{j=1}^{K}
        i_j 2^{j-1}
    \right),
    \qquad
    i_j \in \{0, 1\}.
\end{equation}
Then, a temporal CNN approximates this tensor as a sum
of rank 1 tensors.
The optimal approximation error is hence a consequence of
an Eckart-Young type theorem for higher-order singular value decomposition (HOSVD)%
~\citep{kolda2009.TensorDecompositionsApplications}.
This motivates the definition of an approximation
space that depends on the tail of the singular value sequence.
Let us now make this more precise.
We fix a CNN of depth $K$,
and consider the tensorisation $\tens{\seq{\rho}_{[0,2^K]}}$.
In the theory of HOSVD~\citep{kolda2009.TensorDecompositionsApplications},
this tensor has $2K$ singular values
\begin{equation}\label{eq:hosvs}
    \sigma_1^{(K)}\ge\sigma_2^{(K)}\ge\cdots\ge\sigma_{2K}^{(K)}\geq 0,
\end{equation}
the first $K$ of which are equal and redundant.
The last $K$ singular values determine the error
of low rank approximation of this tensor,
much in the same way as ordinary singular value decay rates
determine the accuracy of low rank approximation of matrices.
Thus, we may consider specifying some decay rate $G$ so that
the tail sum of singular values (which corresponds to low rank approximation error)
satisfies
\begin{equation}\label{eq:g_decay}
    \left(\sum_{i=s+K}^{2K}|\sigma_i^{(K)}|^2\right)^{\frac 1 2} \leq c G(s),
\end{equation}
with $G(s) \rightarrow 0$ as $s\rightarrow \infty$.
Now, we can build an approximation space by considering target functional
sequences whose Riesz representation $\seq \rho$ satisfies the following property:
for each $K$, the singular value tail sum of $\tens{\seq\rho_{[0,2^K]}}$
has a decay rate of at least $G$ (\cref{eq:g_decay}).
Then, the error of low rank approximation of these functional sequences
can be described by $G$.
This leads to the definition of a complexity measure in~\citet{jiang2021.ApproximationTheoryConvolutional}
of the form
\begin{equation}\label{eq:cnn_complexity}
    \begin{aligned}
        C_1(\bm H) =
        \inf \Bigg\{c: \left(\sum_{i=s+K}^{2K}|\sigma_i^{(K)}|^2\right)^{\frac 1 2} \leq c G(s),
        ~s\geq 0, K\ge 1\Bigg\},
    \end{aligned}
\end{equation}
and $G$ is a specified rate of decay of the singular values.
The Jackson-type rate estimate in \cref{eq:jackson_cnn} then follows
from the fact that the maximum rank of a CNN with $K$ layers and $M$ channels is at least $KM^{1/K}$.
The class of functional sequences where $C_1$ is finite defines an approximation
space (with respect to $G$) of sufficiently regular functional sequences
that admits efficient approximation by temporal CNNs.
This is analogous to the characterization of classical smoothness spaces by
the decay rate of series coefficients,
e.g. wavelet coefficients~\citep{mallat2009.WaveletTourSignal}.
Here, we can understand $C_1(\seq{H})$ as a measure of how easy
it is to approximate $\seq{H}$ by tensor product-sums.
In particular, it can be shown~\citep{jiang2021.ApproximationTheoryConvolutional}
that if $\seq{H}$ has a sparse Riesz representation (memory),
then it has small $C_1(\seq{H})$.
This supports the empirical observation that temporal CNNs
excel in applications such as text-to-speech~\citep{oord2016.WaveNetGenerativeModela},
where such sparsity patterns are expected.
}

Let us now contrast this insight to that obtained for the RNN,
which excel at modelling memory patterns that are exponentially
decreasing, but not necessarily sparse.
Consider a target functional sequence with Riesz representation
\begin{equation}
    \rho(t)= \delta(t-2^{K_0}) =
    \begin{cases}
        1 & t = 2^{K_0}, \\
        0 & t \neq 2^{K_0}.
    \end{cases}
\end{equation}
This corresponds to a shift operation, where the output is the result of
shifting the input by $2^{K_0}$ units.
Observe that this target functional sequence
is inside the temporal CNN hypothesis space,
hence it can be exactly represented by setting $K = K_0$ and $M=1$.
However, when $K_0$ is large, it becomes increasingly difficult for a
power sum
$
	u(t) = c_0 + \sum_{i=1}^m c_i\  \gamma_i^{ t}
$
to approximate this function.
The form of $u$ here is a simplified discrete
analogue of the exponential sum
in \cref{eq:hlrnn}.
For any such $u$,
we have the following property
due to~\citet{erdelyi1996.SharpBernsteintypeInequality},
\rev{\begin{equation}\label{eq:exp_property}
    m
    \geq
    \frac{t}{2 \sup_{s\in [0,2t+2]} u(s)}
    |{u(t+1)-u(t)}|.
\end{equation}}%
Since $\rho(t)$ has a sudden change at $t = 2^{K_0}$,
$u(t)$ need at least $2^{K_0-1}$ terms to achieve approximation,
making it challenging for a RNN to learn this target.
Conversely, there exists targets which are easily approximated
(in fact, exactly represented) by $\Hrnn$ but have high complexity
when approximated by $\Hcnn$~\citep{jiang2021.ApproximationTheoryConvolutional}.
These Jackson-type results highlight the interesting
differences between the RNN and the CNN architectures
with respect to the types of sequential relationships
they are adapted to approximating.

\subsection{Encoder-decoder architectures}
\label{sec:others_encdec}


Encoder-decoder architectures%
~\citep{cho2014.LearningPhraseRepresentations,
    cho2014.PropertiesNeuralMachine,
    sutskever2014.SequenceSequenceLearninga,
    kalchbrenner2013.RecurrentContinuousTranslationb}
are a class of sequence to sequence models where
an encoder first maps the input sequence into a fixed-sized context vector,
and then a decoder maps the context vector into the output sequence.
The development of encoder-decoder models was motivated by the need
to handle input and output sequences with varying lengths.
The encoder-decoder architecture is flexible and allows
for the use of various configurations for the encoder and decoder components.

We consider the simplest setting where the encoder and the decoder are both
recurrent networks~\citep{cho2014.LearningPhraseRepresentations,sutskever2014.SequenceSequenceLearninga}.
This has the advantage that we can compare the results here
with those in ordinary RNNs in~\cref{sec:rnn}.
The RNN encoder-decoder ({\rencdec}) architecture
(ignoring bias for simplicity) can be written as
\begin{equation}\label{eq:reddynamics}
    \begin{aligned}
    \dot h(s) &= \sigma_E(Wh(s)+Ux(s)),
    & v &= Qh_0, \quad s\leq 0\\
    \dot g(t) &= \sigma_D(Vg(t)),
    & g_0 &= Pv, \\
    \h{y}(t)  &= c^\top g(t),
    & t &\ge 0,
    \end{aligned}
\end{equation}
where $W\in\R^{m\times m}$, $U\in\R^{m\times d}$, $Q\in\R^{N\times m}$, $V\in\R^{m\times m}$,
$P\in\R^{m\times N}$ and $c\in\R^{m}$.
Sequences $\seq h$ and $\seq g$ are the RNN-type hidden states
corresponding to the encoder and decoder dynamics, respectively.
The encoder is first applied to the entire input sequence $\seq{x}$
in order to produce a fixed-size context vector $v$,
which is the final hidden state of the encoder.
This context vector summarizes the input sequence.
The vector is then utilized as the initial state of the decoder,
which generates an output at each time step.
This defines sequential relationship between two semi-infinite sequences,
with the input $\seq{x}$ having support in $(-\infty, 0]$
and the output $\seq{y}$ having support in $[0, \infty)$.
The complexity of these maps (approximation budget)
is controlled by the RNN width $m$ and context vector size $N$.

Approximation properties of the {\rencdec} architecture are
investigated in detail for the linear case
($\sigma_E,\sigma_D$ are identity maps) in~\citet{li2022.ApproximationPropertiesRecurrent}.
For simplicity of presentation, we take $d=1$,
corresponding to scalar input sequences.
Then, one can rewrite the {\rencdec} hypothesis space as
\begin{equation}\label{eq:hred}
    \Hedl =
    \bigcup_{m, N}
    \Hedl^{(m,N)} =
    \bigcup_{m, N}
    \Big\{\seq{\h{H}}:
    \h{H}_t(\seq x)
    =
    \int_0^\infty \sum_{n=1}^N \h{\psi}_n(t)\h{\phi}_n(s) x(-s)ds
    \Big\},
\end{equation}
where one may recall that
$m$ is the width of the RNNs used for the
encoder and the decoder,
and $N$ is the size of the context vector.
The sequences $\seq{\h \psi_n}$ and $\seq{\h \phi_n}$
are in exponential sum forms
\begin{equation}\label{eq:phi_psi}
    \h\psi_n(t)
    =
    \Bigg(\sum_{i,j=1}^{m}c_i P_{jn}
    \left[e^{Vt}\right]_{ij}\Bigg),
    \qquad
    \h\phi_n(t)
    =
    \Bigg(\sum_{i,j=1}^{m}u_i Q_{nj}
    \left[e^{Wt}\right]_{ji}\Bigg).
\end{equation}
Since the {\rencdec} architecture maps sequences of disjoint support,
it is no longer meaningful to consider time-homogeneity and causality.
Indeed, causality is always satisfied and time-homogeneity is not satisfied.
This is the case by design: the {\rencdec} architecture is used to model
sequential relationships without the shift-equivariant condition.
Consequently, the target functional sequences considered here are
only assumed to be continuous and linear.
In this case, the Riesz representation of these targets take the form
\begin{equation}\label{eq:riesz_encdec_target}
    H_t(\seq{x}) = \int_0^\infty \rho(t,s)^\top x(-s) ds,
    \qquad
    t \geq 0.
\end{equation}
This is a more general form where $\seq \rho$ depends on
two temporal indices $t$ (outputs) and $s$ (inputs) simultaneously.

\begin{figure}[H]
    \centering
        \includegraphics[draft=false,width=0.8\textwidth]{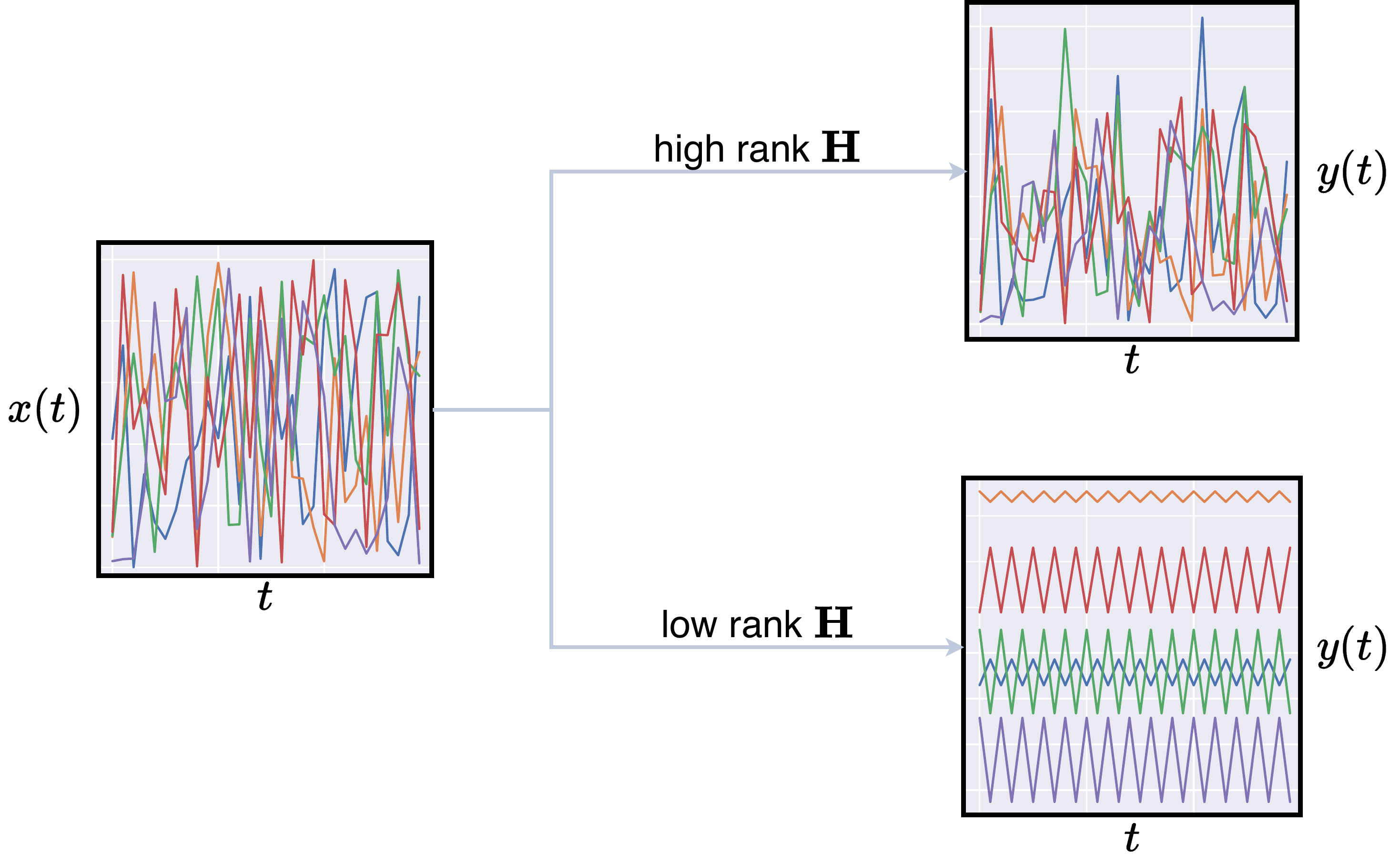}
    \hfill
    \caption{
        Schematic illustration of a high-rank vs low-rank
        sequential relationship under the temporal product structure.
        A dataset of input sequences (left)
        are fed into a functional sequence producing
        the corresponding output sequences (right).
        The top (resp. bottom) right plot shows the resulting sequence
        of a high-rank (resp. low-rank) relationship.
        Observe that the high rank relationship yields a
        complex and input-sensitive temporal structure.
        In contrast, the outputs of the low rank relationship
        exhibit greater regularity, with only macroscopic structures present.
        It is precisely the latter that {\rencdec} is adapted
        to model.
    }
    \label{fig:high_low_rank}
\end{figure}

The density of the hypothesis space~\labelcref{eq:hred} in the space of
sufficiently regular continuous linear functional is established
in~\citet{li2022.ApproximationPropertiesRecurrent}.
This result follows from the observation that we now seek approximations
of $\rho(t,s)$ via a product of two exponential sums.
Hence, one may follow essentially the same approach as in the RNN case
to prove density.
More interestingly, Jackson-type estimates can also be derived.
In particular, one has the following approximation rate
under similar settings as in~\cref{eq:jackson_rnn}
\begin{align}\label{eq:jackson_rencdec}
    \| \seq{H} - \seq{\h{H}} \|
    \leq
    \frac{C_1(\alpha) \gamma}{\beta^2 m^\alpha}
    +
    C_2(\seq{H}, N),
\end{align}
where the meaning of various constants are defined similarly as
in~\cref{eq:jackson_rnn}.
Observe that the first term is similar to the RNN rate
\labelcref{eq:jackson_rnn}, as both the encoder and decoder are implemented
using RNNs.
The estimate $C_2(\seq H, N)$
highlights the new complexity measure associated with
encoder-decoder architectures,
since $N$ is the complexity of the context (coding) vector that acts as the only
intermediary between the encoder and decoder components.
To see what the complexity measure may be,
let us compare~\cref{eq:phi_psi}
and~\cref{eq:riesz_encdec_target}.
Observe that approximating a target $\seq{H}$ simply amounts
to approximating its Riesz representation $\seq{\rho}$ by
a tensor-product summation of the form
\begin{equation}\label{eq:model_svd}
    \h{\rho}(t,s)
    =
    \sum_{n=1}^N \h{\psi}_n(t)\h{\phi}_n(s).
\end{equation}
One may immediately notice that this is a rank-$N$ approximation
of a two-variable function as sums of products of univariate functions.
The optimal approximation is obtained through the proper orthogonal decomposition
(POD)~\citep{chatterjee2000.IntroductionProperOrthogonala},
which is an infinite-dimensional version of the optimal low rank approximation
of matrices via truncated singular value decomposition.
In fact, we may write the formal POD expansion for $\seq{\rho}$ as
\begin{equation}\label{eq:target_svd}
    {\rho}(t,s) = \sum_{n=1}^\infty \sigma_n{\psi}_n(t){\phi}_n(s).
\end{equation}
where $\sigma_n$ are the singular values and $\psi_n,\phi_n$
are the left and right singular vectors (functions).
This is called a \emph{temporal product} structure
in~\citet{li2022.ApproximationPropertiesRecurrent}.
Then, an analogue of the classical Eckart-Young theorem implies
that the optimal approximation error is simply the tail-sum of the singular
values. This is precisely the estimate in $C_2$, i.e.
\begin{equation}
    C_2(\seq{H}, N)
    \propto
    \left(\sum_{n=N+1}^\infty \sigma_n^2\right)^{\frac 1 2}.
\end{equation}
This is considered as the effective rank of the target,
and the Jackson-type estimate in~\cref{eq:jackson_rencdec}
says that a target with low effective rank can be approximated
efficiently with {\rencdec} (with small context vector).
Note that this notion of rank is different from the
tensorisation rank discussed for CNNs in~\cref{sec:others_cnn}.
The concept of effective rank of the sequential relationship
under temporal products is similar to that in linear algebra,
where the rank of a matrix is the dimension of its range space.
This definition can be extended to apply to sequential relationships.
\Cref{fig:high_low_rank} illustrates this idea:
A low rank temporal relationship results in a more
regular output sequence.
In particular, local perturbations to the input sequence result into global
perturbations of the output sequence.
This is very different from both the CNN and the RNN architectures,
and the Jackson-type estimate makes this difference precise.

Currently known approximation results only focus on linear RNN
encoder-decoder.
However, the density result can be extended to non-linear cases by following
the same approaches outlined in~\cref{sec:rnn},
due to the similarity with RNNs.
The rate estimate is less straightforward to extend to non-linear activations.
Nevertheless, one may expect that the uncovered relationship
between the size the context vector and a low-rank type of approximation should
hold generally for encoder-decoder architectures.  This is because in all such
structures, the input and output sequence (both may be infinite-dimensional) only
communicate through a bottle-neck coding vector (finite dimensional), and thus
the approximation should be viewed as a generalized low-rank approximation.

\subsection{Attention-based architectures}
\label{sec:others_attn}


In the final part of this section, we discuss approximation theory
for the growingly popular attention-based architectures.
The attention mechanism was first proposed in~\citet{bahdanau2016.NeuralMachineTranslation}
in the context of RNNs.
Subsequently, it was employed in a variety of practical network architectures.
The attention mechanism, much like the encoder-decoder mechanism,
is a component that can be incorporated into existing models.
Since its introduction, the attention mechanism has become popular tool in applications,
including natural language processing~\citep{vaswani2017.AttentionAllYou}
and computer vision~\citep{dosovitskiy2021.ImageWorth16x16}.
In fact, one of the most successful successful model families,
the Transformer~\citep{vaswani2017.AttentionAllYou},
is based on both the attention mechanism and the encoder-decoder mechanism.
However, our theoretical understanding of the attention mechanism is currently limited,
particularly with regard to its approximation properties.

Let us focus our discussion on the Transformer family of
attention-based architectures.
Currently established approximation results include
the universal approximation capabilities of Transformer
networks~\citep{yun2020.AreTransformersUniversala}
and its sparse variants~\citep{yun2020.ConnectionsAreExpressivea}.
It is important to note that in this context,
the term ``Transformer'' refers specifically to the encoder component
of the original architecture proposed in~\citet{vaswani2017.AttentionAllYou}.

In order to study the Transformer under the sequence approximation setting,
it is convenient to restrict the index set $\set{T}$ to a finite set
$\set{T} = \{1,2,\dots,\tau\}$.
Then, the approximation problem becomes finite-dimensional.
The reason is as follows.
The use of position encoding in Transformer networks is necessary to eliminate
their permutation equivariance (we will show this exactly later).
Position encoding is a sequence $\seq e$
where $t\mapsto e(t)$ is a fixed or trainable
function, independent of $\seq{x}$.
\rev{The sequence $\seq e$ preserves the information of temporal order.}
For training convenience, the length of this encoding is fixed.
As a result, Transformer networks are unable to directly process
infinite-length sequences,
unlike RNNs and CNNs based architectures.

The simplest transformer block consists of the following components,
\begin{equation}\label{eq:transformer_block}
    \begin{aligned}
    \text{Attn}(\seq x)(t) &= x(t) +  \sum_{i=1}^q
    W_o^{i} \sum_{s=1}^\tau\sigma [(W_{Q}^{i}x(t))^\top W_{K}^{i}x(s)] \, W_{V}^{i} x(s),\\
    \text{Trans}(\seq x)(t) &= \text{Attn}(\seq x)(t) + f(\text{Attn}(\seq x)(t)),
\end{aligned}
\end{equation}
where
$
    W_{Q}^{i}, W_{K}^{i}, W_{K}^{i}
    \in
    \R^{m\times d}$, $W_{o}^{i} \in \R^{d\times m}
$.
Here, $\text{Attn}(\seq x)$ is the attention block,
$\sigma$ is a normalization usually taken as the softmax function,
and $\tau$ is the maximum input sequence length.
The attention mechanism produces an output which is
subsequently fed into a common trainable feed-forward network $f$,
pointwise in time.
This constitutes a Transformer block.
Define the Transformer hypothesis space by
\begin{equation}
    \Htrans^{(m_1, m_2, q, l)}
    =
    \Big\{\seq{\h{H}}:
    \seq{\h{H}}
    \text{ is a composition of $l$ Transformer blocks } t^{(m_1,m_2,q)}
    \Big\},
\end{equation}
where $t^{(m_1,m_2,q)}=\text{Trans}(\cdot)$ is a Transformer block defined
in \labelcref{eq:transformer_block},
$m_1$ is the trainable dimension of the attention block
(total degrees of freedom of $W_{o}, W_{Q}, W_{K}, W_{V}$),
$q$ is the number of attention heads
and $m_2$ is the dimension of the trainable parameters in the
pointwise feed-forward network $f$.
In~\citet{yun2020.ConnectionsAreExpressivea}, a sparse variants is defined,
where the $W_Q^i$ matrix in the attention block satisfies certain sparsity conditions.
We denote the sparse Transformer hypothesis space by
\begin{equation}
    \Htransp^{\,(m_1, m_2, q, l)} \subset \Htrans^{\,(m_1, m_2, q, l)},
\end{equation}
which is a subset of the Transformer hypothesis space.

We start with density results for the Transformer.
First, note that without position encoding,
the Transformer hypothesis space is permutation equivariant.
Concretely, let $p$ be a permutation of the sequence index,
which is a bijection on $\{1,\dots,\tau\}$.
For a sequence $\seq{x}$, we denote by $\seq{x}\circ p$
the permuted sequence $[\seq{x}\circ p](t) = x(p(t))$.
A functional sequence $\bm H$ is said to be
permutation equivariant if for all $p$ and $\bm x$
we have $\bm H(x \circ p) = \bm H(\bm x) \circ p$.
We can check that the Transformer block~\labelcref{eq:transformer_block}
is permutation equivariant if one does not perform positional encoding.
This certainly limits approximation properties,
and thus we hereafter assume that
a fixed position encoding is added to the input $\seq x$,
such that the model input becomes $\seq x + \seq e$.

In~\citet{yun2020.AreTransformersUniversala,yun2020.ConnectionsAreExpressivea},
a density results for the Transformer is proved under the following conditions.
Assuming the target $\seq H$ is continuous, and the input sequence space
is uniformly bounded,
then $\seq{H}$ can be approximated by
\begin{equation}
    \seq{\h H} \in \bigcup_l
    \Htransp^{( 1, 4 ,2, l)}\subset \bigcup_l  \Htrans^{( 1, 4 ,2, l)}.
\end{equation}
This result is proved by a special construction.
First, one uses a stack of attention blocks to achieve the following condition:
\begin{enumerate}
  \item For any input $\seq{x}$,
  the value of the output $\seq{\tilde{x}}$ are all distinct.
  \item For all inputs $\seq{x_1}, \seq{x_2}$ such that
  $\seq{x_1} \neq \seq{x_2}$,
  their outputs $\seq{\tilde{x}_1}$ and $\seq{\tilde{x}_2}$ have no common value.
\end{enumerate}
These conditions can be understood as for each $t$,
$\tilde{x}(t)$ captures the information of the entire input sequence.
Next, a deep stack of pointwise feed-forward blocks
are constructed to map each $\tilde{x}(t)$ to the desired output.
This construction results in a deep Transformer architecture
with a small width.

However, this construction is not generally how Transformer operates,
since the first part of the construction is an attention-only network,
which is shown to degenerate quickly ~\citep{dong2021.AttentionNotAlla}.
In a similar vein,
several studies such as~\citet{cordonnier2020.RelationshipSelfAttentionConvolutional}
and~\citet{li2021.CanVisionTransformersa}
have demonstrated that a Transformer can represent
a CNN through careful parameterization.
Therefore, density results from CNN imply
the density of the transformer.
Again, there is little empirical evidence that the Transformer
behaves like a CNN in applications.

The ability for the Transformer to mimic other architectures
is not surprising, since it has many highly flexible
components (encoder-decoders, fully connected networks, attention mechanisms)
that can be carefully, but often artificially, adjusted to represent
other known architectures as a special case.

In fact, we give here another example of such a representation
that, to the best of our knowledge,
has not been reported in the literature,
but is straight-forward to derive.
We can show that a two-layer Transformer can mimic the form of
a generalized Kolmogorov representation theorem~\citep{ostrand1965.DimensionMetricSpaces}.
This result states that for $d$ dimensional compact sets $I_1,\dots,I_\tau \subset \mathbb R^d$,
any continuous
$\seq H : I_1 \times \cdots \times I_\tau \to \R$
can be written as
\begin{equation}
    H(\seq x) = \sum_{q=0}^{2d\tau}
    \Phi_q \left(  \sum_{s=1}^\tau  \phi_{q,s}(x(s))\right),
\end{equation}
where $\Phi_q$ and $\{\phi_{q,s}\}$ are continuous functions.
It is possible to design a two-layer Transformer
exhibiting a similar form, implying density.
For simplicity, we consider $d=1$ and only the output at $H_t$ when $t = 1$.
The general case can be constructed similarly.
One can ensure that with position encoding,
a pointwise feed-forward function is able to apply different
mappings at each temporal index position.
To see this, observe that for a collection of continuous functions
$f_i: \mathbb [0,1]^d \to \mathbb R$,
$i=1,\dots,\tau$,
we can find vectors $\{e_i\}$
and a continuous function $F:\R^d\to\R$
such that
$
    F(x + e_i) = f_i(x).
$
Now suppose we have an input sequence $\seq x$.
\begin{itemize}
    \item Layer 1.
    We set $W_o^i = 0$ in the attention block,
    so that the input directly goes into the pointwise feed-forward block.
    From the previous discussion,
    the pointwise feed-forward network
    can be constructed to give an output
    $y^{(1)} :\R \to \R^{2\tau + 1}$,
    such that
    $
            y^{(1)}(s) =
            c_s[
                \h{\phi}_{0,s}(x(s)),
                \dots,
                \h{\phi}_{2\tau,s}(x(s))
            ]^\top
    $,
    with $c_1 = 1/2$ and $c_s=1$ when $s>1$.
    Due to the density of feed-forward neural networks,
    each $\h{\phi}_{j,s}$ can be chosen to approximate any continuous function.
    \item Layer 2.
    In the attention block, by letting $W^i_{K}=0$,
    the softmax function gives a constant output where
    $\text{softmax} [(W_{Q}^{i}x(t))^\top W_{K}^{i}x(s)]\equiv \frac{1}{\tau}$.
    Let $W_{V}^i = I, W_o^i = \tau I$ and $h = 1$, then we have
    $
        \text{Attn}(\seq y^{(1)})(1) = \sum_{s=1}^\tau  y^{(1)}(s).
    $
    Hence, the final output after the feed-forward network
    with linear readout $c^\top = (1,\dots, 1) \in \R^{2\tau+1}$ gives
    \begin{equation}
        \begin{aligned}
            H_1(\seq x)
            &=
            c^\top \h{\Phi}(\text{Attn}(\seq y^{(1)})(1)), \\
            &=
            \sum_{q=0}^{2\tau}\h{\Phi}_q\left(\sum_{s=1}^{\tau} [y^{(1)}(s)]_q\right), \\
            &=
            \sum_{q=0}^{2\tau}\h{\Phi}_q\left(\sum_{s=1}^{\tau} \h{\phi}_{q,s}(s)\right),
        \end{aligned}
    \end{equation}
    where $\h{\Phi}$ is again a feed-forward neural network
    that can be adjusted to approximate any continuous function.
    Thus, the Kolmogorov representation can be approximated through
    this particular construction.
\end{itemize}

This highlights a common issue in current approximation results for
complex structures such as the transformer.
Density-type results are rarely illuminating, since they can be
constructed in many ways due to the structural flexibility.
However, they rarely reveal the working principles of the complex
model under study.
In particular, it gives little insights to why and when these
models should be used for applications.
Such insights may result from finer analysis of approximation properties,
including Jackson-type and Bernstein-type results as outlined
for the other architectures previously.


\revtwo{
To date, there are few - if any - Jackson or Berstein-type
results for sequence modelling using the Transformer.
We mention a related series of works on static function approximation
with a variant of the Transformer architecture%
~\citep{kratsios2022.SmallTransformersCompute,kratsios2022.UniversalApproximationConstraints,acciaio2022.DesigningUniversalCausal}.
Here, the targets are continuous functions $H: [0, 1]^\tau \to K$,
and $K\subset \R^n$ is a compact set.
Examples include classification problems
where $K$ is a probability simplex, and
covariance matrix prediction problems with $K$ being
the set of symmetric positive semi-definite matrices.
The authors consider a variant of the transformer architecture
to approximate this target function family.
For $x \in \mathbb [0,1]^\tau$, and $Y_1, \cdots, Y_N \in K$,
an approximant of the following form is considered,
\begin{equation}
    \begin{aligned}
        \h H(x) &= \text{Attn}(\h D(\h E (x)),Y) \\
        &= \sum_{i=1}^N \text{softmax}(\h D(\h E (x)))_{i} \delta_{Y_i},
    \end{aligned}
\end{equation}
where
$\h E: \mathbb R^\tau \to \mathbb R^m$,
$\h D: \mathbb R^m \to \mathbb R^N$ are two fully-connected neural networks,
and $\delta_{Y_i}$ is a point mass at $Y_i$.
The softmax is taken along the $i=1,\dots,N$ direction.
One may observe the deviations of this architecture
compared with the Transformer used
in sequence modelling \labelcref{eq:transformer_block}.
For this modified architecture,
the authors derive an approximation error estimate
based on increasing the complexities of the encoder $\h E$ and the decoder $\h D$.
The identified notion of regularity for the target
to induce efficient approximation is smoothness,
similar to classical approximation of functions.
However, the sequence approximation properties of
the practical Transformer architecture \labelcref{eq:transformer_block},
and in particular its relation to memory structures in the data,
remains an open problem.
This is an important direction of future research.
}

%% file: outlook.tex
\section{Discussion and outlook}
\label{sec:outlook}


Let us first summarize in~\cref{tab:summary}
the approximation results we discussed
in~\crefrange{sec:rnn}{sec:others}.
Observe that most results in the literature
are of the density-type,
and current Jackson and Bernstein-type results
are often limited to the simplified setting of linear activations.
Nevertheless, these rate estimates are instructive
in revealing some key insights on the approximation
of sequence relationships using different architectures.

\begin{table}[ht!]
    \begin{center}
    \caption{Summary of approximation results for sequence modelling.}
    \label{tab:summary}
    \begin{tabular}{l|ccc}
        \toprule
        & \textbf{Density-type}
        & \textbf{Jackson-type}
        & \textbf{Bernstein-type}\\
        \midrule
        \textbf{RNN} & \checkmark & \rev{Barron, linear} & linear\\
        \textbf{CNN} & \checkmark & linear & -\\
        \textbf{\rencdec} & \checkmark & linear & - \\
        \textbf{Transformer} & \checkmark & - & -\\
        \bottomrule
    \end{tabular}
\end{center}
\end{table}

We can collectively summarize this insight as
a form of \emph{structural compatibility}.
That is to say:
\begin{verse}
    \emph{
        Each model architecture is efficient in approximating
        precisely those targets that resemble its temporal structure.
    }
\end{verse}
For example, we saw that RNNs are particularly good at
approximating relationships with an exponentially decaying memory pattern.
We can attribute this to the fact that RNNs themselves have
an exponentially decaying memory structure, as evidenced
by the expression~\cref{eq:hlrnn}.
Similarly, temporal CNNs are effective in approximating
targets whose memory structure has low-rank under tensorisation,
i.e. can be written as the product-sum of few tensors.
\rev{This is indeed what the temporal CNN itself looks like:
we recall that the tensorisation rank of the temporal CNN
with $K$ layers and $M$ channels scales like $K M^{1/K}$.}
The same holds for the RNN encoder-decoder with respect
to its low-rank structure under temporal products,
induced by the context vector.
\rev{
We emphasize that this notion of rank is very different from that
in temporal CNN approximation.
In the convolution case, the rank refers to
the tensorisation procedure related to the stacked
convolutional structure of the temporal CNN.
In the case of recurrent encoder-decoders,
the rank refers to the amount of temporal coupling between the input
sequence and the output sequence.
Mathematically, this coupling is measured by the
rate of decay of the singular values in the expansion
\labelcref{eq:target_svd}.
The recurrent encoder-decoder with small context vector
precisely parameterize a temporal relationship that
has little coupling between inputs and outputs.
The Jackson-type results show that the recurrent
encoder-decoder is indeed adapted to approximate
target relationships having the same property.
}
The requirement of structural compatibility is consistent
with classical approximation theory.
For example, trigonometric polynomials with low-orders
are themselves smooth functions with small gradient-norms,
and thus are adapted to approximating these functions.
The same pattern is observed for non-linear approximation,
wavelets and multi-resolution analysis, where weakened smoothness,
sparsity and multi-scale structures dictate both the model
structures and effective targets for their application~\citep{devore1998.NonlinearApproximation}.

Now, let us discuss future research directions to further
our understanding of the approximation theory of sequence modelling.
Besides the obvious task of completing~\cref{tab:summary},
we may wish to ask:
\emph{What does a successful
theory of sequence approximation entail?}
While there is no singular definition of success,
it is reasonable to discuss desired outcomes
in two broad categories.

On the practical side,
one pressing need is to reduce the amount of
trial and error during model selection.
The understanding of the suitability of different model
architectures for different problem types
is essential in guiding implementations in practice.
Therefore, an important task is to formalize
a model selection workflow for sequence modelling.
This certainly requires more than approximation theory,
but the current understanding already suggests that
we should quantify the memory patterns observed in
datasets to select the model archetype.
Developing this concrete pipeline based on
well-understood theory is of great interest
and importance.
Another practical application worth noting
is the simplification of model architectures.
Modern architectures developed for specific
applications may be very complex,
and it is likely that some of their components
are not performance critical.
A theory of sequence modelling should help
to identify the components that may not be
necessary, so as to simplify and distil
the essential modelling techniques.


On the mathematical side,
following the development of classical approximation
theory~\citep{devore1998.NonlinearApproximation},
it is of interest to characterize the so-called approximation
spaces that are associated with each sequence modelling hypothesis space.
Recall that the results for RNN in~\cref{sec:rnn}
suggests a type of approximation space in the form of
\begin{equation}
    \Crnn
    =
    \{
        \seq{H} \in \set{C}
        :
        \normCrnn{\seq H} < \infty
    \},
\end{equation}
where the norm $\normCrnn{\cdot}$ may take the form
\begin{equation}
    \normCrnn{\seq H}
    =
    \|\seq{H}\|
    +
    |\seq{H}|_*.
\end{equation}
Here, $\|\seq{H}\|$ is the usual uniform norm
$\|\seq{H}\| = \sup_{t}\sup_{\|\seq{x}\|_{L^\infty}\leq 1} |H_t(\seq{x})|$,
and $|\seq{H}|_*$ is a suitable semi-norm
measuring exponential decay.
For example, motivated by \cref{eq:jackson_rnn},
we may take
\begin{equation}
    |\seq{H}|_*
    =
    \sup_{t\in\R}
    \max_{r=1,\dots,\alpha+1}
    \sup_{\seq{x} \in \set{X}_0}
    |e^{\beta t} H_t^{(r)}(\seq{x})|
\end{equation}
where $\beta$ is the supremum value for which
$|e^{\beta t} H_t^{(r)}(\seq{x}) |$
remains finite for all $t$, $\seq{x} \in \set{X}_0$,
and $r=1,\dots,\alpha+1$.
The set $\set{X}_0\subset\set{X}$
is a suitable set of test sequences.
One can check that $(\Crnn, \normCrnn{\cdot})$
forms a normed linear subspace.
Then, the Jackson-type estimate can be
rewritten as the familiar form
\begin{equation}
    \inf_{\seq{\h H} \in \Hrnnl^m}
    \| \seq{H} - \seq{\h H} \|
    \leq
    \text{Constant }
    \times
    \frac{
        \normCrnn{\seq H}
    }{
        m^\alpha
    }.
\end{equation}
Hence, this suggests that the approximation space $\Crnn$
is the RNN analogue of the usual Sobolev spaces
($\mathcal{W}^{\alpha,\infty}$) that characterizes
trigonometric polynomial approximation.
The space $\Crnn$ is reminiscent of the
Schwartz spaces~\citep{stein2011.FourierAnalysisIntroduction}
arising in Fourier analysis,
except that we are now concerned with exponentially
(instead of polynomially) decaying derivatives.
\rev{
    Note that here, we are primarily concerned with
    the effect of temporal structure on approximation.
    Since time is one-dimensional, regularity related
    to smoothness should be expected.
    In the case where the ambient dimension of the
    input sequence $d$ is large (and the relationship is non-linear),
    one expects that smoothness alone is insufficient
    to ensure efficient approximation.
    In this case, one may envision approximation spaces
    with a combination of smoothness conditions in the temporal direction
    and Barron-type conditions in the spatial direction.
}
Similar constructions of approximation spaces
can be made from Jackson-type results for the other architectures
we described before.
To characterize these spaces, their interpolation theory
and whether they correspond to familiar spaces arising from
analysis is of keen mathematical interest.
Another aspect is characterizing the difference
between linear and non-linear approximation.
Taking the RNN as an example,
the usual RNN (with trainable $W,U,b$) is a
non-linear hypothesis space, in the sense that
the linear combination of two functional sequences from $\Hrnn^m$
is in general a new functional sequence not in $\Hrnn^m$,
but $\Hrnn^{2m}$.
On the contrary, reservoir computing systems
take $W,U,b$ as fixed random realizations,
and $\Hesn^m \oplus \Hesn^m = \Hesn^m$.
That is to say, the $\Hesn^m$ is a linear approximation
space.
In classical approximation theory,
linear and non-linear (adaptive) approximation
lead to different approximation spaces%
~\citep{devore1998.NonlinearApproximation}.
It is thus of interest to investigate this distinction
for sequence modelling, e.g.,
clarifying the difference of using ESNs versus RNNs
for approximation.

Beyond approximation theory,
it is important to note that a comprehensive understanding
of sequence modelling should also account for
optimization and generalization aspects.
Indeed, principled sequence modelling in machine learning
is not only the design of model architectures,
but also how to train them
and how to regularize them to maximize
testing performance.
For example, it is observed that while
RNN training can be shown to be stable
in the certain regimes%
~\citep{hardt2018.GradientDescentLearns,allen-zhu2019.ConvergenceRateTraining},
it can sometimes be provably ineffective
in the presence of long-term memory%
~\citep{li2021.CurseMemoryRecurrent,li2022.ApproximationOptimizationTheorya}.
Generalization theories have also been explored,
e.g. in
\citet{chen2019.GeneralizationBoundsFamily,tu2020.UnderstandingGeneralizationRecurrent,wang2021.ProvableGeneralizationRecurrent}.
\revtwo{
Other than deterministic model families,
there are also results on modeling sequential data via
(latent) neural controlled (stochastic) differential equations,
such as hybrid architectures with GANs~\citep{kidger2021.NeuralSdesInfinitedimensional},
universal neural operators for causality~\citep{galimberti2022.DesigningUniversalCausal},
and neural SPDE models motivated by
mild solutions~\citep{hu2022.NeuralOperatorRegularity,salvi2021.NeuralStochasticPDEs}.
Applications include time series generation~\citep{lozano2023.NeuralSDEsConditional},
irregular and long time series analysis~\citep{kidger2020.NeuralControlledDifferential,morrill2021.NeuralRoughDifferential},
and online prediction~\citep{morrill2022.ChoiceInterpolationScheme}.
These interesting aspects of sequence modelling theory
are beyond the scope of the current survey.
}